\documentclass[runningheads]{llncs}
\usepackage{graphicx}
\usepackage{wrapfig}
\usepackage{tikz}
\usepackage{comment}
\usepackage{amsmath,amssymb} 
\usepackage{color}
\usepackage{multirow}
\usepackage{kotex}
\usepackage{subcaption}
\usepackage{adjustbox}
\usepackage{float}

\usepackage{booktabs}  

\usepackage{xcolor}
\newcommand{\myparagraph}[1]{\vspace{3pt}\noindent{\bf #1}}

\newcommand{\etal}{{\it{et al.}}}

\begin{document}
\pagestyle{headings}
\mainmatter
\def\ECCVSubNumber{5584}  

\title{Sound-guided Semantic Video Generation} 

\titlerunning{Sound-guided Semantic Video Generation}
%
\authorrunning{Lee, Seung Hyun et al.}

\author{Seung Hyun Lee\inst{1}\index{Lee, Seung Hyun} \and
Gyeongrok Oh\inst{1}\and
Wonmin Byeon\inst{2} \and
Chanyoung Kim\inst{1} \and \\
Won Jeong Ryoo\inst{1}\index{Ryoo, Won Jeong} \and
Sang Ho Yoon\inst{3}\index{Yoon, Sang Ho} \and
Hyunjun Cho\inst{1} \and 
Jihyun Bae\inst{1} \and \\
Jinkyu Kim\inst{4*} \and
Sangpil Kim\inst{1*}
}

%
%
\institute{
$^1$Department of Artificial Intelligence, Korea University\\
$^2$NVIDIA Research, NVIDIA Corporation \\ 
$^3$Graduate School of Culture Technology, KAIST\\
$^4$Department of Computer Science and Engineering, Korea University \\
}

\maketitle

\begin{abstract}
The recent success in StyleGAN demonstrates that pre-trained StyleGAN latent space is useful for realistic video generation. However, the generated motion in the video is usually not semantically meaningful due to the difficulty of determining the direction and magnitude in the StyleGAN latent space. 
In this paper, we propose a framework to generate realistic videos by leveraging multimodal (sound-image-text) embedding space. As sound provides the temporal contexts of the scene, our framework learns to generate a video that is semantically consistent with sound.
First, our sound inversion module maps the audio directly into the StyleGAN latent space. We then incorporate the CLIP-based multimodal embedding space to further provide the audio-visual relationships. Finally, the proposed frame generator learns to find the trajectory in the latent space which is coherent with the corresponding sound and generates a video in a hierarchical manner. 
We provide the new high-resolution landscape video dataset (audio-visual pair) for the sound-guided video generation task. The experiments show that our model outperforms the state-of-the-art methods in terms of video quality. We further show several applications including image and video editing to verify the effectiveness of our method. 

\end{abstract}
\section{Introduction}

Existing video generation methods rely on motion generation from a noise vector given an initial frame~\cite{tian2021a,tulyakov2018mocogan,yan2021videogpt}. As they create trajectory from noise vectors without any guidance, the motion of the generated video is not semantically meaningful. On the other hand, sound provides the cue for motion and various context of the scene~\cite{9524590}. Specifically, sound may represent events of the scene such as `viola playing' or `birds singing'. It can also provide a tone of the scene such as `Screaming' or `Laughing'.  This is an important cue for generating a video because the video's temporal component and motion are closely associated with sound. Generating high-fidelity video from sound is crucial for a variety of applications, such as multimedia content creation and filmmaking. However, most sound-based video generation works focus on generating a talking face matching from verbal sound using 2D facial landmarks~\cite{chen2019hierarchical,Das2020SpeechDrivenFA,Suwajanakorn2017SynthesizingO} or 3D face representation~\cite{Chen2020TalkingheadGW,Richard_2021_WACV,Thies2020NeuralVP,wang2021facevid2vid,wu2021imitating,yi2020audiodriven,Yang:2020:MakeItTalk}.

\let\thefootnote\relax\footnote{\scriptsize{{\bf Acknowledgement.} This work is partially supported by Institute of Information \& communications Technology Planning \& Evaluation (IITP) grant funded by the Korea government(MSIT) (No. 2019-0-00079, Artificial Intelligence Graduate School Program(Korea University)).
 J. Kim is partially supported by the National Research Foundation of Korea grant (NRF-2021R1C1C1009608), Basic Science Research Program (NRF-2021R1A6A1A13044830), and ICT Creative Consilience program (IITP-2022-2022-0-01819). Any opinions, findings, and conclusions or recommendations expressed in this material are those of the authors and do not necessarily reflect the views of the funding agency.}}
\let\thefootnote\relax\footnotetext{\scriptsize{$^*$Corresponding authors: Jinkyu Kim and Sangpil Kim}. Code and more diverse examples are available at \url{https://kuai-lab.github.io/eccv2022sound/}.}
%
Generating a realistic video from non-verbal sound is challenging. First, the spatial and temporal contexts of a generated video need to be semantically consistent with sound. Although some studies~\cite{chatterjee2020sound2sight,le2021ccvs} have tried to synthesize video with sound as an independent variable, mapping from 1D semantic information from audio to visual signal is not straightforward. Furthermore, when a sound of waves enters as an input, these works can not generate a video with diverse waves. Second, the generated motion should be physically plausible and temporally coherent between frames. Recent works~\cite{brouwer2020audio,jeong2021tr} use a latent mapper network with music as the input for navigating in StyleGAN~\cite{karras2019style} latent space. Since the direction in the latent space is randomly provided, however, the content semantics from the video are not realistic. Finally, there is a lack of high-fidelity video dataset for generating a realistic video from non-verbal sound. 

\begin{figure}[t]
    \centering
    \includegraphics[width=\linewidth]{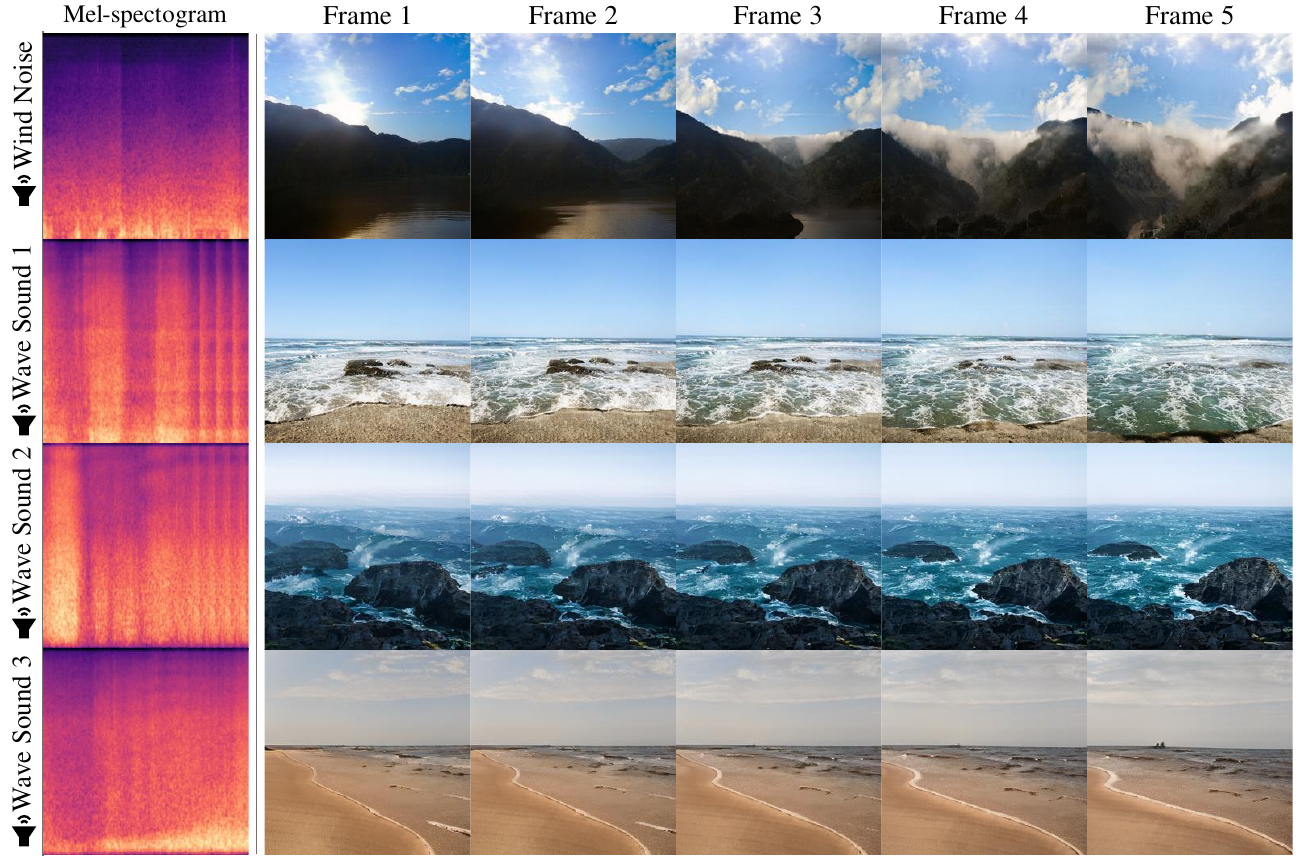}
    \caption{An example of sound-guided semantic video generation. Our method generates semantic video reflecting temporal context using audio as an input feature.
    }
    \label{fig:video-example}
\end{figure}

To overcome these challenges, we propose a novel framework that can semantically invert sound to the StyleGAN latent space for video generation. 
The difficulty of inverting audio into the StyleGAN latent space is that the image regenerated after inverting the audio is visually irrelevant to the audio. To resolve this issue, we employ the joint embedding prior knowledge learned from large-scale multimodal data (image, text, sound) to sound inverting. 
A high fidelity video is generated by moving the latent vector in the $\mathcal{W}+$ space which is disentangled latent space of StyleGAN. Our sound-encoder learns to put the latent space $\mathcal{W}+$ from the sound input and find the trajectory from the initial latent code for video generation. As shown in Fig.~\ref{fig:video-example}, we generate a video whose meaning is consistent with the given sound-input. 

\begin{wrapfigure}{r}{0.5\textwidth}
    \centering
    \includegraphics[width=0.5\textwidth]{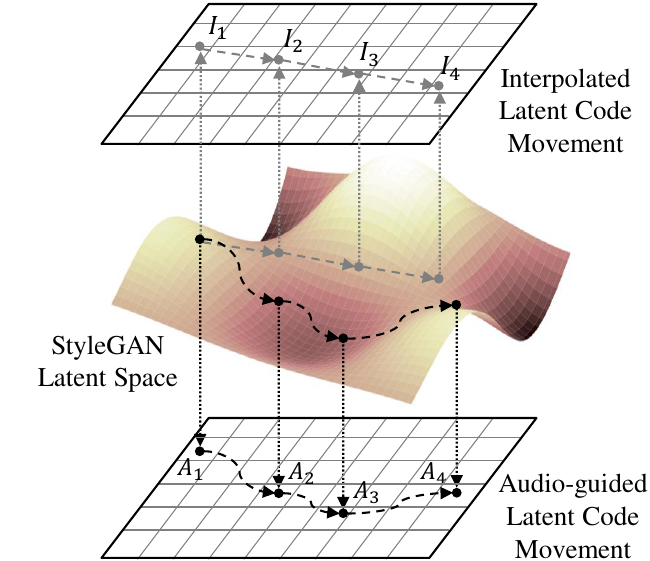}
    \caption{
    Comparison of traversing method for the video generation. The top grid shows the conventional interpolated way, and the bottom grid shows our audio-guided latent code movement method on StyleGAN's latent space.
    }
    \label{fig:interpolation-Analysis}
\end{wrapfigure}

Contrastive Language-Image Pre-Training~(CLIP)~\cite{radford2learning} creates a very powerful joint multimodal embedding space with 400 million image-text pair data. We leverage a CLIP-based multimodal embedding space~(sound-image, text-image) trained on a large scale at the same time. This multimodal embedding space provides guidance when walking in the $\mathcal{W}+$ space. StyleCLIP~\cite{Patashnik_2021_ICCV} and TediGAN~\cite{xia2021tedigan} leverage the representational power of CLIP for text-driven image editing. Recently, the prior knowledge of CLIP has been transferred to the audio-visual relationship, making it possible to develop various applications using CLIP in sound modality~\cite{wu2021wav2clip,lee2021sound}. In this paper, we exploit the prior knowledge of audio-visual to generate a complete video related to sound. Furthermore, we introduce a frame generator that finds the desired trajectory from the sound. The frame generator predicts the latent code of the next time by using audio as a constraint. StyleGAN's style latent vector has a hierarchical representation, and our frame generator which is divided into three layers predicts coarse, mid, and fine style, respectively. Therefore, our proposed method provides nonlinear guidance for each time step beyond the linear interpolation of latent vectors~(see Fig.~\ref{fig:interpolation-Analysis}). The latent code of our proposed method travels in the guidance of the audio in the latent space of StyleGAN with generating audio semantic meaningful and diverse output. In contrast, the general interpolation method travels in the latent space in one direction without any guide at regular intervals.


In this paper, our experimental results demonstrate that the proposed method supports a variety of sound sources and generates video that is a better reflection of given audio information.
The existing audio-visual benchmark datasets~\cite{chen2020vggsound,kay2017kinetics,Kuehne11} have limitations in high-fidelity video generation. We release a new high-resolution landscape video dataset, as the existing datasets are not designed for high-fidelity video generation~\cite{chen2020vggsound} or do not support sound-video pairs~\cite{kay2017kinetics,Kuehne11}.
Our dataset is effective for experimenting with high-resolution audio-visual generation models. 

Our main contributions are listed as follows:
\begin{itemize}
    \item We propose a novel method for mapping audio semantics into the StyleGAN latent space by matching multi-modal semantics with CLIP space. 
    \item We introduce a framework for semantic-level video generation solely based on the given audio. We demonstrate the effectiveness of the proposed method by outperforming the quality of generated videos from state-of-the-art methods in the sound-guided video generation task.
    \item Our framework not only determines the direction and size of movement of the latent code in the StyleGAN latent space, but also can generate rich style video given various sound attributes. 
    \item We provide the new high-resolution landscape video dataset with audio-visual pairs for this task.
\end{itemize}

\section{Related Work}
\myparagraph{StyleGAN Latent Space Analysis.} 

StyleGAN~\cite{karras2019style} has a semantically rich latent space, so many studies~\cite{Patashnik_2021_ICCV,richardson2021encoding,xia2021tedigan} have analyzed the latent space of GAN. Most of the existing StyleGAN inversion modules minimize the difference between a latent vector and an embedding vector that encodes an image generated from the vector. 
Richardson~\textit{et al.}~\cite{richardson2021encoding} have effectively inverted high-quality face images using prior knowledge related to the faces such as LPIPS~\cite{zhang2018perceptual}, perceptual loss and pretrained ArcFace~\cite{deng2019arcface}. Patashnik~\textit{et al.}~\cite{Patashnik_2021_ICCV} defines a style latent mapper network that projects a given latent code to match the meaning of the text prompt. The latent mapper of the study learns given a pair of image and text prompts sampled from noise. However, there is a limitation in that randomly sampled image cannot have continuous motion like a real video.
Our goal is to indicate the direction in which the style latent code will move in an audio-visual multimodal embedding space. Our study embeds the input audio directly into the StyleGAN latent space. The sound-guided latent code enables video generation in the expanded audio-visual domain. CLIP~\cite{radford2learning} represents the powerful joint embedding space between image and text modalities with 400 million image-text pairs, and this joint embedding space provides direction for traversing the StyleGAN latent space.
Recent studies~\cite{guzhov2021audioclip,lee2021sound,wu2021wav2clip} extend the modalities of CLIP to audio. Lee~\textit{et al.}~\cite{lee2021sound} especially focused on audio-visual representation learning for image editing, and we also leverage that audio-visual multimodal space embedding for navigating the latent code.
Our works differ from previous works in that we consider temporal style variation guided by the CLIP embedding space. Sound semantics generate images at each time step using the latent vector optimized by the CLIP guidance. 

\myparagraph{Sound-guided Video Generation.}  
Many studies use the temporal dynamics in sound as a source for vivid video generation~\cite{chatterjee2020sound2sight,jeong2021tr,le2021ccvs}. 
There are mainly two approaches to generate video from sound as a source. The first is a conditional variational autoencoder method~\cite{kingma2013auto} that predicts the distribution of future video frames in the latent space. VAE-based method~\cite{chatterjee2020sound2sight,yan2021videogpt,le2021ccvs} mainly solve video prediction tasks that predict the next frame for a given frame. Among them, Chatterjee,~\textit{et al.}~\cite{chatterjee2020sound2sight} and Le,~\textit{et al.} \cite{le2021ccvs} consider the context of a scene with a non-verbal sound. Ji~\textit{et al.}~\cite{ji2021audio} and Lahiri~\textit{et al.}~\cite{lahiri2021lipsync3d} generate a video about facial expression with a verbal sound.

In this study, we enable the video generation task with a non-verbal sound condition that is more difficult than the verbal sound condition. GAN-based video generation is to sample the video from a noise vector~\cite{saito2017temporal,tulyakov2018mocogan,vondrick2016generating,wang2020g3an}.
Tian~\textit{et al.}~\cite{tian2021a} and Fox~\textit{et al.}~\cite{fox2021stylevideogan} synthesize continuous videos with StyleGAN~\cite{karras2019style}, a pre-trained high-fidelity image generator. 
Unlike previous studies, we consider a novel sound-guided high resolution video generation method. Jeong~\textit{et al.}~\cite{jeong2021tr} explores StyleGAN's latent space to generate a video. However, the domain of the sound is limited to music, and the guidance is not noticed by the user. 
We generate high-resolution video considering the semantics of sounds such as wind, raining, etc.
\section{Sound-guided Video Generation}


\begin{figure}[t]
    \centering
    \includegraphics[width=\linewidth]{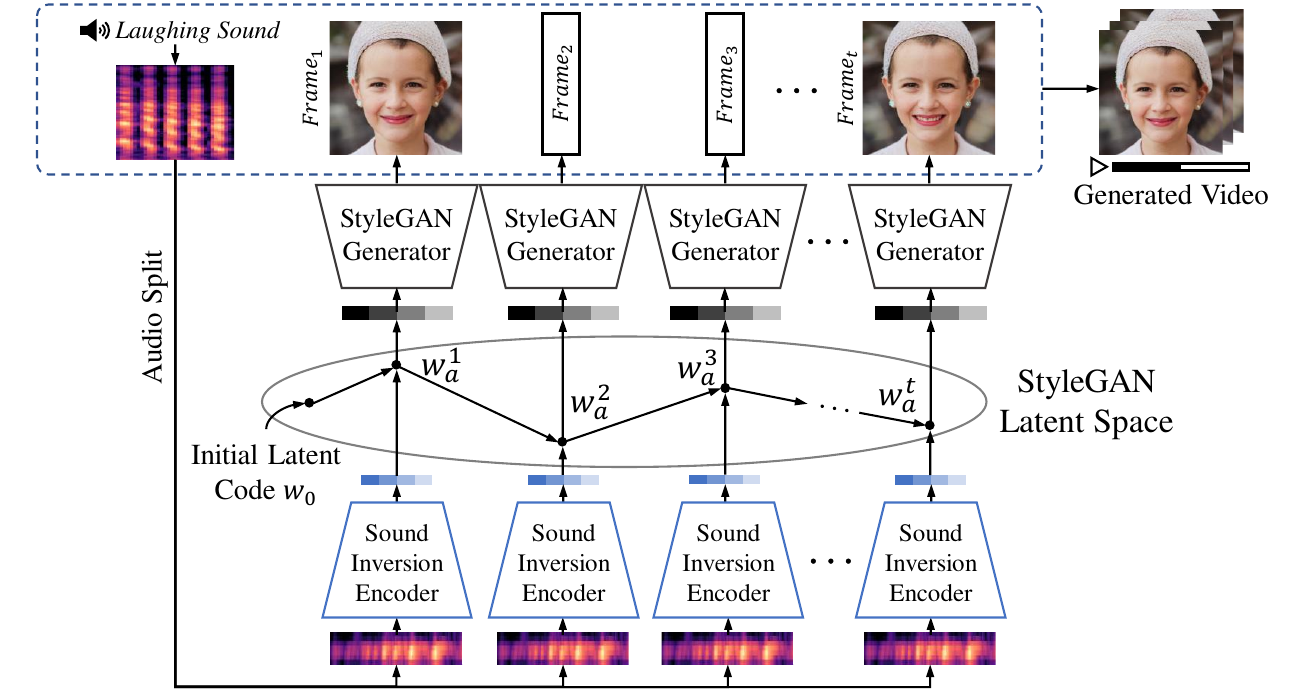}
    \caption{An overview of our proposed sound-guided video generation model. Our model consists of two main modules: (i) Sound Inversion Encoder (Section~\ref{sec:sound-inversion-encoder}), which takes a sequence of audio inputs as an input and outputs a latent code to generate video frames. (ii) StyleGAN-based Video Generator (Section~\ref{sec:video-generator}), which generates temporally consistent and perceptually realistic video frames conditioned on the sound input.
    }
    \label{fig:overview}
\end{figure}

Our model takes sound information as an input to generate a sequence of video frames accordingly, as shown in Fig.~\ref{fig:overview}. For example, given a Laughing sound input, our model generates a video with a facial expression of laughing. To achieve this goal, our model needs two main capabilities. (i) The ability to understand the sound input and to condition it in the trained video generator. (ii) The ability to generate a video sequence that is 
perceptually realistic and is temporally consistent. 

We propose that such capabilities can be learned via our two novel modules: (1) {\em Sound Inversion Encoder}, which learns a mapping from the sound input to a latent code in the (pre-trained) StyleGAN~\cite{karras2019style} latent space (Section~\ref{sec:sound-inversion-encoder}). (2) {\em StyleGAN-based Video Generator}, which is conditioned on the sound-guided latent code and generates video frames accordingly (Section~\ref{sec:video-generator}). Furthermore, we leverage the representation power of the CLIP~\cite{radford2learning}-based multimodal (image, text, and audio) joint embedding space, which regularizes perceptual consistency between the sound input and the generated video.

\begin{wrapfigure}{r}{0.5\textwidth}
\centering
\includegraphics[width=0.5\textwidth]{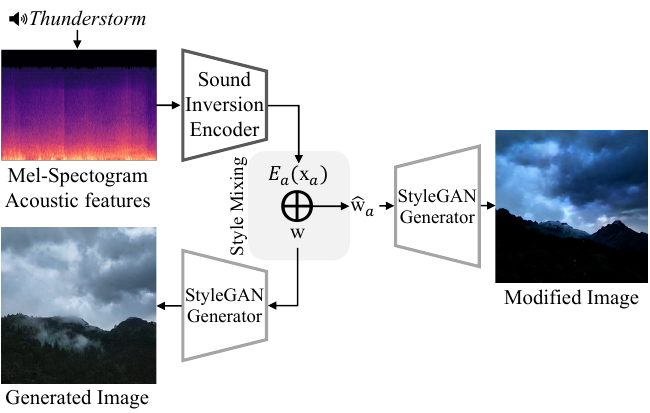}
\caption{
Audio Inversion Encoder. From a given sound input, sound-guided latent code is obtained by the elementwise summation with the randomly sampled latent code.
}
\label{fig:inversion_module}
\end{wrapfigure}


\subsection{Inverting Sound Into the StyleGAN Latent Space}\label{sec:sound-inversion-encoder}
 
As shown in Fig.~\ref{fig:inversion_module}, our Sound Inversion Encoder takes as an input Mel-spectrogram Acoustic features and outputs a latent feature ${\bf{w}}_a\in\mathcal{W}+$ in the pre-trained StyleGAN feature space $\mathcal{W}+$. The sound-conditioned latent feature ${\bf{w}}_a$ is then augmented by the element-wise summation with the randomly sampled latent code ${\bf{w}}\in\mathcal{W}+$, yielding a sound-guided latent code ${\hat{\bf{w}}_a}\in\mathcal{W}+$. Conditioned on ${\hat{\bf{w}}}_a$, we generate an image, which maintains the content of the original image (generated with the random latent code ${\bf{w}}$) but its style is transferred according to the semantic of sound. Formally,
\begin{equation}
{\bf{\hat{w}}}_a = E_a({\bf{x}}_a) + {\bf{w}}.
\label{loss:image_manipulation}
\end{equation}
where $E_a(\cdot)$ denotes our Sound Inversion Encoder given a sound input ${\bf{x}}_a$.

\myparagraph{Matching Multimodal Semantics via CLIP Space.}
Lee~\etal~\cite{lee2021sound} introduced an extended CLIP-based multi-modal (image, text, and audio) feature space, which is trained to produce a joint embedding space where a positive triplet pair (e.g., audio input: “thunderstorms”, text: “thunderstorm”, and corresponding image) are mapped close together in the CLIP-based embedding space, while pushing that of negative pair samples further away. We utilize this pre-trained CLIP-based embedding space to generate images that are semantically well-aligned with the sound input. Specifically, we minimized the following cosine distance (in the CLIP embedding space) between the image generated from the latent code ${\hat{\bf{w}}_a}$ and an audio input ${\bf{x}}_a$.
    \begin{equation}
    \mathcal{L}_\text{CLIP}^{(a \leftrightarrow v)} = 1 - {F_v(G({\bf{\hat{w}}}_a))\cdot F_a({\bf{x}}_a) \over ||F_v(G({\bf{\hat{w}}}_a))||_2 \cdot ||F_a({\bf{x}}_a)||_2}.
    \label{loss:image_generation_clip}
\end{equation}
where $G(\cdot)$ denotes the StyleGAN~\cite{karras2019style} generator, while $F_v(\cdot)$ and $F_a(\cdot)$ are CLIP's image encoder and audio encoder, respectively. A similar loss function could be used for a pair of audio and text prompts. CLIP's text embedding provide constraints on the semantics of audio chunks.
Using audio labels (e.g. thunderstorm) as a text prompt, we minimize the cosine distance between representations for image and text in the CLIP embedding space. 

Lastly, we use $l_2$ distance between the latent code ${\hat{\bf{w}}_a}$ and their averaged latent code, i.e. ${\bar{\bf{w}}_a}=\sum_t {\hat{\bf{w}}_a^t}$, to explicitly constrain the generated sequence of images to share similar contents and semantics over time. Ultimately, we minimize the following loss to train our Sound Inversion Encoders. 

\begin{equation}
\mathcal{L}_\text{enc}=\mathcal{L}_\text{CLIP}^{(a \leftrightarrow v)} + \mathcal{L}_\text{CLIP}^{(a \leftrightarrow t)} + \lambda_b||{\bf{\hat{w}}}_a - \bar{{\bf{w}}}_a||^2_2.
\label{loss:image_generation_total}
\end{equation}
where $\lambda_b$ controls the strength of the regularization term.

\begin{figure}[t]
    \centering
    \includegraphics[width=\linewidth]{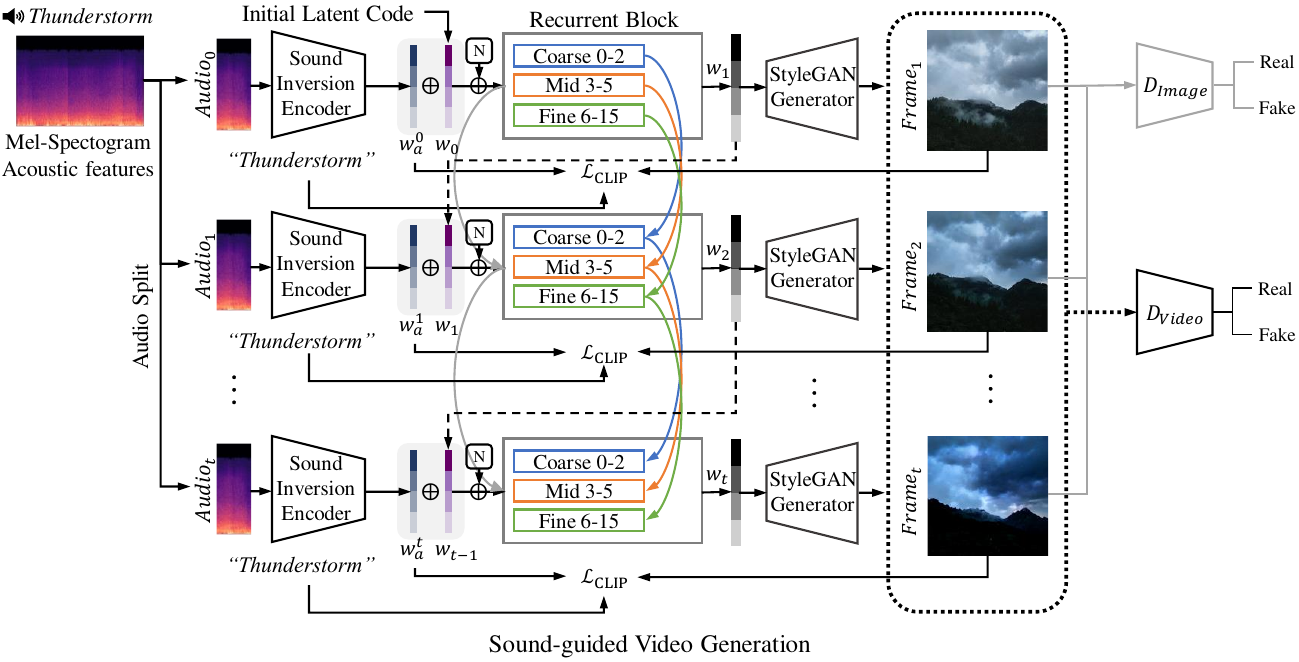}
    \caption{An overview of our sound-guided video generation model, which consists of two main parts: (i) Sound Inversion Encoder, which iteratively generates sound-conditioned latent code ${\bf{w}}_a^t$ from corresponding audio time segments. (ii) StyleGAN-based Video Generator, which recurrently generates a video frame that is trained to be consistent with neighboring frames. Moreover, we train image and video discriminators adversarially to generate perceptually realistic video frames.}
    \label{fig:video-generation}
\end{figure}   

\subsection{Sound-guided Semantic Video Generation}\label{sec:video-generator}
\myparagraph{Recurrent Module for Latent Code Sequence Generation.}
As shown in Fig.~\ref{fig:video-generation}, our model recurrently generates latent code at each timestep to generate an image sequence with the StyleGAN~\cite{karras2019style} generator. Formally, we train a recurrent neural network $E_{\text{RNN}}(\cdot)$ that outputs ${\bf{\hat w}}_a^t$ conditioned on the previous audio segment ${\bf{x}}^{t-1}_a$ and the hidden state ${\bf{h}}_t$ at timestep $t$.
\begin{equation}
    {\bf \hat{w}}_a^t, {\bf h}_t = E_\text{RNN}({\bf h}_{t-1}, E_a({\bf{x}}_a^{t-1}) + {\bf{\hat{w}}}^{t-1}_a),
    \label{loss:transformer_input}
\end{equation}
where $E_{\text{RNN}}(\cdot)$ comprises a fully-connected layers to recurrently yield the hidden state ${\bf{h}}_t$ and the current latent code ${\hat{\bf w}}_a^{t}$.

\myparagraph{Multiple Recurrent Blocks for Rich Style Information.}
The input latent vectors are divided into three groups (0-2, 3-5, and 6-15 layers) and fed to different recurrent networks to generate different levels of detail.
These three latent groups represent different styles (coarse, mid, and fine features) and control the attributes in StyleGAN~\cite{Patashnik_2021_ICCV}. 
Each recurrent block predicts the latent code of the next time step:
\begin{equation}
    {\bf{\hat{w}}}_a^t = (E_\text{RNN}^\text{coarse}({\bf{\hat{w}}}_{a}^{t-1}), E_\text{RNN}^\text{mid}({\bf{\hat{w}}}_{a}^{t-1}), E_\text{RNN}^\text{fine}({\bf{\hat{w}}}_{a}^{t-1})),
    \label{loss:transformer_coarse}
\end{equation}
where $E_\text{RNN}^\text{coarse}(\cdot)$, $E_\text{RNN}^\text{mid}(\cdot)$, and $E_\text{RNN}^\text{fine}(\cdot)$ denote coarse, mid, and fine recurrent encoder-decoder network, respectively. 
Generated video $\tilde{\bf{v}}$ is a sequence of images synthesized from each latent code as follows:
\begin{equation}
    \tilde{\bf{v}} = [G({\bf{\hat{w}}}_a^1), G({\bf{\hat{w}}}_a^2),...,G({\bf{\hat{w}}}_a^T)].
    \label{loss:transformer_output}
\end{equation}
where $T$ is the total sequence length of the video.

\myparagraph{Adversarial Image and Video Discriminators.} 
As shown in Fig.~\ref{fig:video-generation}, we train a video discriminator $D_V$ adversarially by forwarding the generated video into the discriminator, which determines whether the input video is real or synthesized. Specifically, following MoCoGAN-HD~\cite{tian2021a}, our video discriminator is based on the architecture of PatchGAN~\cite{pix2pix2017}. An input video (i.e. a real or fake example) is divided into small 3D patches, which are then classified as real or fake. The average response is used as the final output. We thus use the following adversarial loss $\mathcal{L}_{D_V}$:
\begin{equation}
    \mathcal{L}_{D_V} = \mathbb{E}[\text{log}D_V({\bf{v}})]+\mathbb{E}[1-\text{log}D_V(\tilde{\bf{v}})],
\end{equation}
where ${\bf{v}}$ and $\tilde{\bf{v}}$ are the real and fake example, respectively. Additionally, we also apply an image discriminator $D_I$, which similarly trains the model adversarially to determine whether the input image is real or fake on the time axis. Concretely, we optimize the following loss function:
\begin{equation}
    \mathcal{L}_{D} = \mathcal{L}_{D_V} + \mathcal{L}_{D_I}  =\mathcal{L}_{D_V} + \mathbb{E}[\text{log}D_I({\bf{v}})]+\mathbb{E}[1-\text{log}D_I(\tilde{\bf{v}})].
    \label{loss:adversarial_loss}
\end{equation}


\myparagraph{Loss Function.} 
Our model can be trained end-to-end by optimizing the following objective:
\begin{equation}
     \underset{\theta_{G}}{\text{min}}\,\underset{\theta_{D_V}}{\text{max}}\,\mathcal{L}_{D_V} + \underset{\theta_{G}}{\text{min}}\,\underset{\theta_{D_I}}{\text{max}}\,\mathcal{L}_{D_I} +\underset{\theta_{E_a}}{\text{min}}\, \lambda_{\text{enc}}\mathcal{L}_{\text{enc}},
\end{equation}
where $\lambda_{\text{enc}}$ controls the strength of the loss term.



\section{Experiments}

\begin{table}[t]
    \caption{Comparison table of the High Fidelity Audio-visual Landscape Video Dataset and other datasets. ($^\dagger$): We report the smallest resolution for comparisons.}
    \centering
    \label{table:dataset-comparison}
    \resizebox{.95\textwidth}{!}{%
    \begin{tabular}{lrrrcr}\toprule
        \parbox{2cm}{\centering Dataset } & \parbox{2cm}{\centering \# of Videos} & \parbox{2cm}{\centering Resolution$^\dagger$} & \parbox{2cm}{\centering \# of Classes} & \parbox{2cm}{\centering Audio-Video Pairs} & \parbox{2cm}{\centering Total Length (hours)} \\\midrule
        HMDB51~\cite{Kuehne11} & $6,766$ & $340 \times 256$ & $51$ & $\times$ & $4.9$ \\
        UCF-101~\cite{soomro2012ucf101} & $13,320$  & $320 \times 240$ & $101$     & $\checkmark$ & $26.7$ \\
        VGG-Sound~\cite{chen2020vggsound} & $200,000$ & $640 \times 360$ & $310$ & $\checkmark$ & $560.0$ \\
        Kinetics-400~\cite{kay2017kinetics} & $306,245$ & $640 \times 360$ & $400$ & $\checkmark$ & $850.7$\\ \midrule
        Sub-URMP~\cite{li2018creating} & $72$  & $1920 \times 1080$& $13$ & $\checkmark$ & $1.0$ \\
        Landscape (ours) & $9,280$   & $1280 \times 720$ & $9$ & $\checkmark$ & $25.8$ \\
        \bottomrule
    \end{tabular}}
\end{table}

\setlength{\tabcolsep}{1.4pt}


\begin{table}[!t]
    \begin{center}
    \caption{Comparison to the state-of-the-art methods. We compare two version of our method (with and without sound inputs) to several state-of-the-art methods. We reproduce Sound2Sight~\cite{chatterjee2020sound2sight}, CCVS~\cite{le2021ccvs}, and Tr\"{a}umerAI~\cite{jeong2021tr} as a baseline on two benchmark datasets: Sub-URMP~\cite{li2018creating} and our created Landscape. 
    }
    \label{table:quantative-sota}
    \resizebox{\textwidth}{!}{
        \begin{tabular}{lcccccc}\toprule
        \multirow{2}{*}{Method} & \multirow{2}{*}{\parbox{2cm}{\centering {\em use} sound inputs}} & \multirow{2}{*}{\parbox{2cm}{\centering {\em use} the first frame}} & \multicolumn{2}{c}{Sub-URMP~\cite{li2018creating}} & \multicolumn{2}{c}{Landscape} \\ \cmidrule{4-7}
        & & & IS~($\uparrow$) & FVD~($\downarrow$) & IS~($\uparrow$) & FVD~($\downarrow$)\\\midrule    
        Sound2Sight~\cite{chatterjee2020sound2sight}  & \checkmark & \checkmark & \hfill 1.64 $\pm$ 0.11 & \hfill 282.48 & \hfill 1.55 $\pm$ 0.48 & \hfill 311.55\\
        CCVS~\cite{le2021ccvs}  & \checkmark & \checkmark & \hfill 2.06 $\pm$ 0.07 & \hfill 274.01 & \hfill 1.78 $\pm$ 0.64 & \hfill 305.40\\
        Tr\"{a}umerAI~\cite{jeong2021tr}   & \checkmark & - & \hfill 1.02 $\pm$ 0.14 & \hfill 350.80 & \hfill 1.07 $\pm$ 0.58 & \hfill 732.63  \\
        \midrule
        Ours ({\em w/o sound inputs})  &  - & - & \hfill 2.99 $\pm$ 0.48 & \hfill 272.27 & \hfill 1.68 $\pm$ 0.43 & \hfill 305.91 \\
        Ours  & \checkmark & - & \hfill \textbf{3.05 $\pm$ 0.42} & \hfill \textbf{271.03} & \hfill \textbf{1.82 $\pm$ 0.26} & \hfill \textbf{291.88} \\\bottomrule
    \end{tabular}}
\end{center}
\end{table}

      \begin{table}[!t]
        \caption{Quantitative evaluation. (a) Semantic Consistency between a given audio and generated video. (b) Ablation study of CLIP Loss. We evaluate the Inversion Score on the LHQ dataset.}
        \begin{subtable}[b]{0.5\textwidth}
            \centering
    
            \caption{Semantic Consistency. }
            \label{table:quantative-consistency}
            \resizebox{\textwidth}{!}{%
            \begin{tabular}{lcc}
            \toprule
            Model & Similarity~(t $\leftrightarrow$ v)$\uparrow$ & Similarity~(a $\leftrightarrow$ v)$\uparrow$\\\midrule
            Sound2Sight~\cite{chatterjee2020sound2sight} & 0.2491 & 0.1771 \\
            CCVS~\cite{le2021ccvs} & 0.2514 & 0.1764 \\
            Tr\"{a}umerAI~\cite{jeong2021tr} & 0.1932 & 0.1416 \\
            Ours & \textbf{0.2556} & \textbf{0.1897} \\
            \bottomrule
            \end{tabular}}
        
        \end{subtable}
        \hfill
        \begin{subtable}[b]{0.5\textwidth}
            \caption{The LHQ Dataset Inversion Score.}
            \label{table:quan_inversion}
            \resizebox{\textwidth}{!}{%
            \begin{tabular}{lcc}
            \toprule
            Model & LPIPS $\downarrow$ & MSE $\downarrow$\\\midrule
            $\mathcal{L}_2 + \mathcal{L}_{reg}$ &  0.648 &  0.052 \\
            $\mathcal{L}_2 + \mathcal{L}_\text{LPIPS} + \mathcal{L}_{reg}$ & 0.468 &  0.070 \\
            $\mathcal{L}_2 + \mathcal{L}_\text{LPIPS} + \mathcal{L}_{reg} + \mathcal{L}_\text{CLIP}$ & \textbf{0.432} & \textbf{0.048}\\
            \bottomrule
        \end{tabular}}
         \end{subtable}
    \end{table}


\subsection{Datasets}
There exist few publicly available Audio-Video paired datasets (e.g. UCF-101~\cite{soomro2012ucf101}, VGG-Sound~\cite{chen2020vggsound}, Kinetics-400~\cite{kay2017kinetics}), but they mostly support low-resolution video and contain a lot of noise in the audio. (see Table~\ref{table:dataset-comparison}). To the best of our knowledge, there is the Sub-URMP (University of Rochester Musical Performance) dataset~\cite{li2018creating} that only provides pairs of high-fidelity audio-video. This dataset provides $72$ video-audio pairs ($\approx$ 1 hour in total) from recordings of $13$ kinds of instruments played by different orchestra musicians. This dataset would be a good starting point to evaluate the performance of video generation models, but it only focuses on orchestra playing scenes with a limited number of video clips, limiting training efficiency. Thus, we create a new dataset that provides a high-fidelity audio-video paired dataset on landscape scenes. 


\myparagraph{High Fidelity Audio-Video Landscape Dataset (Landscape).}
We collect 928 high-resolution (at least 1280 $\times$ 720) video clips, where each clip is divided into 10 (non-overlapped) different clips of 10 seconds each. Overall, the total number of video clips available is 9,280, and the total length is approximately 26 hours. Our Landscape dataset contains 9 different scenes, such as thunderstorms, waterfall burbling, volcano eruption, squishing water, wind noise, fire crackling, raining, underwater bubbling, and splashing water. We provide details of our created Landscape dataset in the supplemental material.

\begin{figure}[t]
    \centering
    \includegraphics[width=\linewidth]{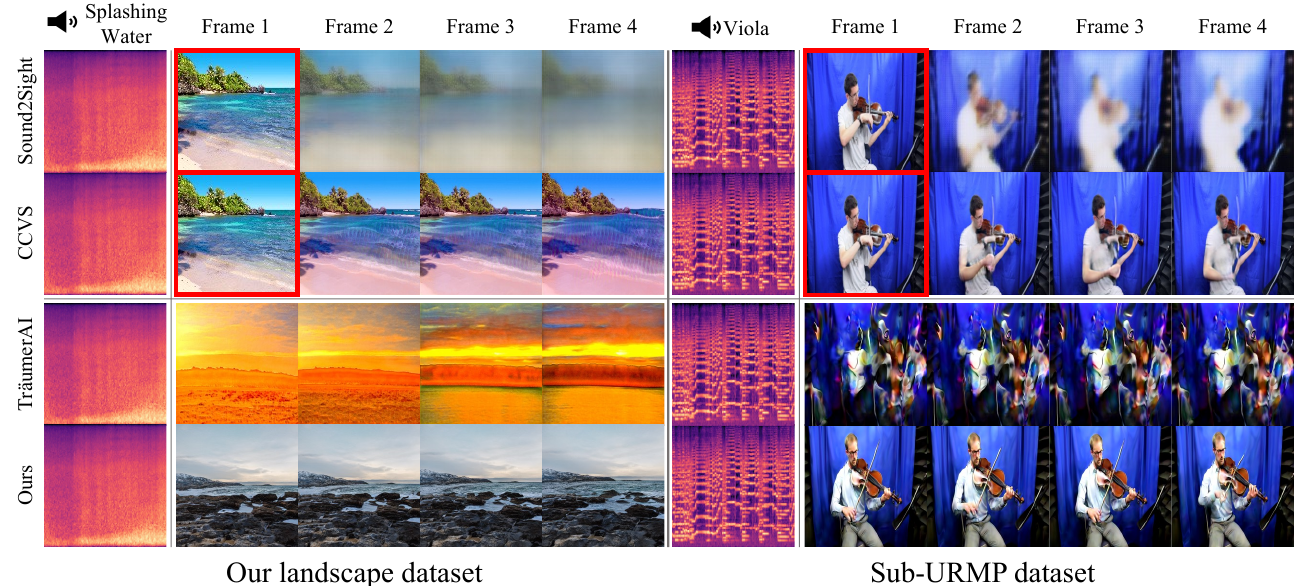}
    \caption{Qualitative comparison between our sound-guided semantic video generation method and previous video generation results on our landscape dataset and Sub-URMP dataset~\cite{li2018creating}. Sound2Sight~\cite{chatterjee2020sound2sight} and CCVS~\cite{le2021ccvs} (top) generate a video by conditioning on an image as the first frame highlighted in red, whereas  Tr\"{a}umerAI~\cite{jeong2021tr}, and our method (bottom) generate a video conditioned on audio input only.
    }
    
    \label{fig:qualitive-comparision-consistency}
\end{figure}

\subsection{Comparison to Existing Sound-guided Video Generation}
\myparagraph{Quantitative Evaluation.} 
We use the latest VAE-based models as baselines, including Sound2Sight~\cite{chatterjee2020sound2sight}, CCVS~\cite{le2021ccvs} and StyleGAN-based model Tr\"{a}umerAI~\cite{jeong2021tr}. 
For our model and Tr\"{a}umerAI, we first pre-train StyleGAN on the high fidelity benchmark datasets~(the Sub-URMP~\cite{li2018creating} and the LHQ datasets~\cite{Skorokhodov_2021_ICCV}) then train to navigate the latent space with the fixed image generator. The videos are generated with randomly sampled initial frames. In contrast, for Sound2Sight and CCVS, the first frames are provided. Table~\ref{table:quantative-sota} shows that our approach produces the best quality results, and sound information is effective for video generation.


Table~\ref{table:quantative-consistency} demonstrates that our method produces more visually correlated videos from sound than other methods. To measure if the generated videos are semantically related to sound, we compare the cosine similarity between text-audio and video embedding. We obtain 512-dimensional video embeddings corresponding to the number of each frame with the CLIP~\cite{radford2learning} image encoder, and average them. Additionally, text embeddings are obtained from CLIP's text encoder, a 512-dimensional vector, and audio embeddings sharing CLIP space are obtained from Lee's~\cite{lee2021sound} multi-modal embedding space. So leveraging the multi-modal embedding space helps to achieve semantic consistency in sound-guided video generation. 

\myparagraph{Evaluation Metrics.} 
Evaluation of the quality of generated videos is known to be challenging. We first use the two widely-used quantitative metrics: Inception Score (IS)~\cite{salimans2016improved} and Fréchet Video Distance (FVD)~\cite{unterthiner2018towards}. The former is widely used to evaluate the outputs of GANs by measuring the KL-divergence between each image's label distribution and the marginal label distribution. Further, FVD quantifies the video quality by measuring the distribution gap between the real vs. synthesized videos in the latent space. We use an Inception3D~\cite{carreira2017quo} network, which is pre-trained on Kinetics-400~\cite{kay2017kinetics} and is fine-tuned on each benchmark accordingly.

\myparagraph{Qualitative Evaluation.} In Fig.~\ref{fig:qualitive-comparision-consistency}, we visually compare the quality of the generated video with other models including Sound2Sight~\cite{chatterjee2020sound2sight}, CCVS~\cite{le2021ccvs} and Tr\"{a}umerAI~\cite{jeong2021tr} for the two benchmark datasets~(Sub-URMP~\cite{li2018creating} and our landscape dataset). 
We found that Tr\"{a}umerAI is not able to produce any realistic video at all. Also, the generated video by Sound2Sight and CCVS is mostly distorted. Our method, on the other hand, produces high fidelity videos.

\subsection{Ablation Studies}
\begin{figure}[t]
    \centering
    \includegraphics[width=\linewidth]{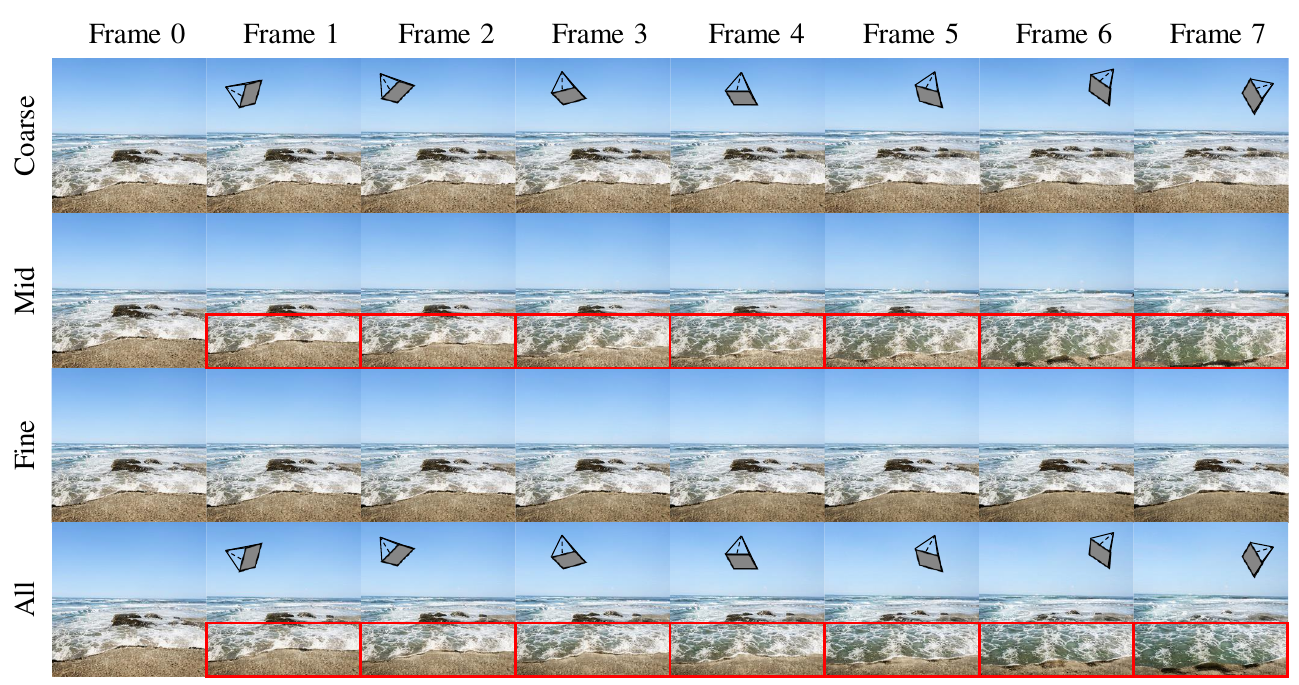}
    \caption{Ablation of multiple recurrent blocks for rich style information. The first row shows the video generation result when only the coarse recurrent block is applied, the second is mid, the third is the fine recurrent block, and the last is the entire block. 
    The red box is the area where the style has changed.
    }
    \label{fig:ablation}
\end{figure}


\myparagraph{CLIP Loss for StyleGAN Inversion.} 
In this study, we show that CLIP~\cite{radford2learning} prior knowledge is helpful for StyleGAN inversion.
Table~\ref{table:quan_inversion} compares the mean squared error~(MSE) and LPIPS~\cite{zhang2018perceptual} between the original and the reconstruction image from the inversion module using the LHQ dataset. By minimizing the cosine distance between CLIP embeddings, the inversion reconstruction performance is improved in landscape images. 

\myparagraph{Effect of Multiple Recurrent Blocks for Rich Style Information.} In sound-guided video generation, multiple recurrent blocks control the StyleGAN attribute. Fig.~\ref{fig:ablation} compares video generation using only one coarse, mid, and fine recurrent block. Each block contains diverse style information. 
Specifically, the coarse and middle recurrent blocks control viewpoint and  semantically meaningful motion changes (such as a wave strike) of the scene. Finally, the fine recurrent block handles fine texture changes.

\subsection{User Study}

In order to evaluate our proposed method, we request one hundred Amazon Mechanical Turk (AMT) participants.
We show participants three types of generated videos that are generated by Sound2Sight~\cite{chatterjee2020sound2sight}, CCVS~\cite{le2021ccvs}, Tr\"{a}umerAI~\cite{jeong2021tr}, our model and ground truth. Participants answer the following questionnaire based on five-point Likert scale: (i) Realness~-~\textit{Please evaluate the realness of the video}~(``1 - very unrealistic'' to ``5 - very realistic'') and (ii) Naturalness~-~\textit{Which video generation result better expresses the target attribute?} (iii) Semantic Consistency~-~\textit{Please evaluate the semantic consistency between video and audio}~(``1 - very inconsistent'' to ``5 - very consistent'').

\begin{figure}[t!]
     \centering
     \begin{subfigure}[b]{0.32\textwidth}
         \centering
         \includegraphics[width=\textwidth]{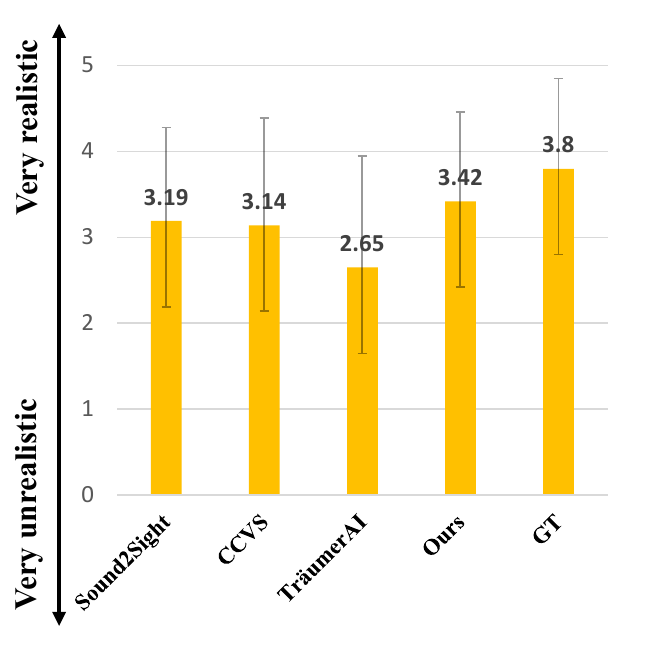}
         \caption{Realness.}
         \label{fig:y equals x}
     \end{subfigure}
     \hfill
     \begin{subfigure}[b]{0.32\textwidth}
         \centering
         \includegraphics[width=\textwidth]{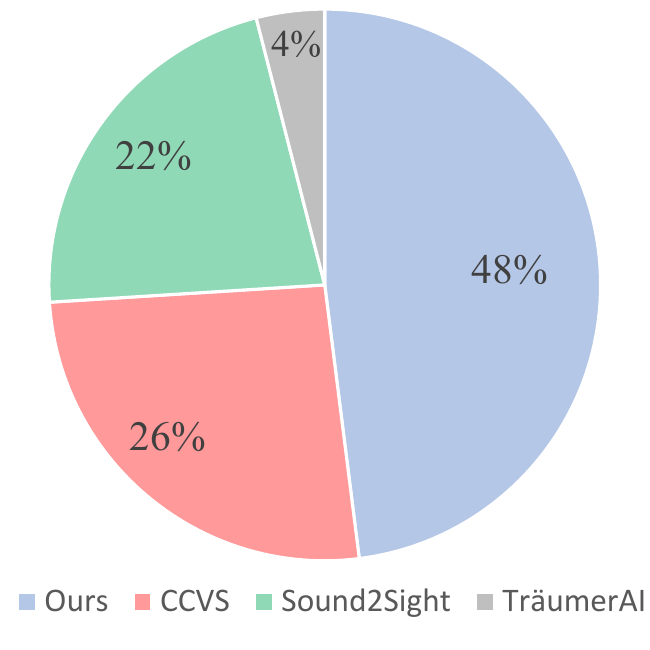}
         \caption{Naturalness.}
         \label{fig:three sin x}
     \end{subfigure}
     \hfill
     \begin{subfigure}[b]{0.32\textwidth}
         \centering
         \includegraphics[width=\textwidth]{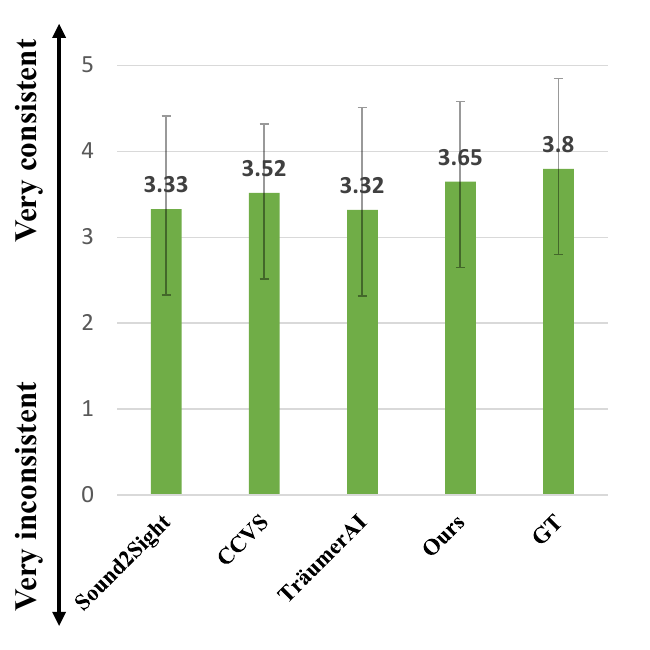}
         \caption{Semantic Consistency.}
         \label{fig:five over x}
     \end{subfigure}
        \caption{User Study Results on our landscape dataset. (a) Realness. (b) Naturalness. (c) Semantic Consistency.}
        \label{fig:three_metric}
\end{figure}
The survey procedure is as follows. First, we have participants watch ground truth videos and measure realness and semantic consistency on a scale of 1 to 5. Then have participants measure realness against the results generated by other models. To evaluate naturalness, we ask participants to choose which of the videos generated by each model best expresses the target attribute.
As shown in Fig.~\ref{fig:three_metric}, our method significantly outperforms other state-of-the-art approaches~(\textit{Realness}, \textit{Naturalness}, \textit{Semantic Consistency}). 

\begin{figure}[t]
    \centering
    \includegraphics[width=\linewidth]{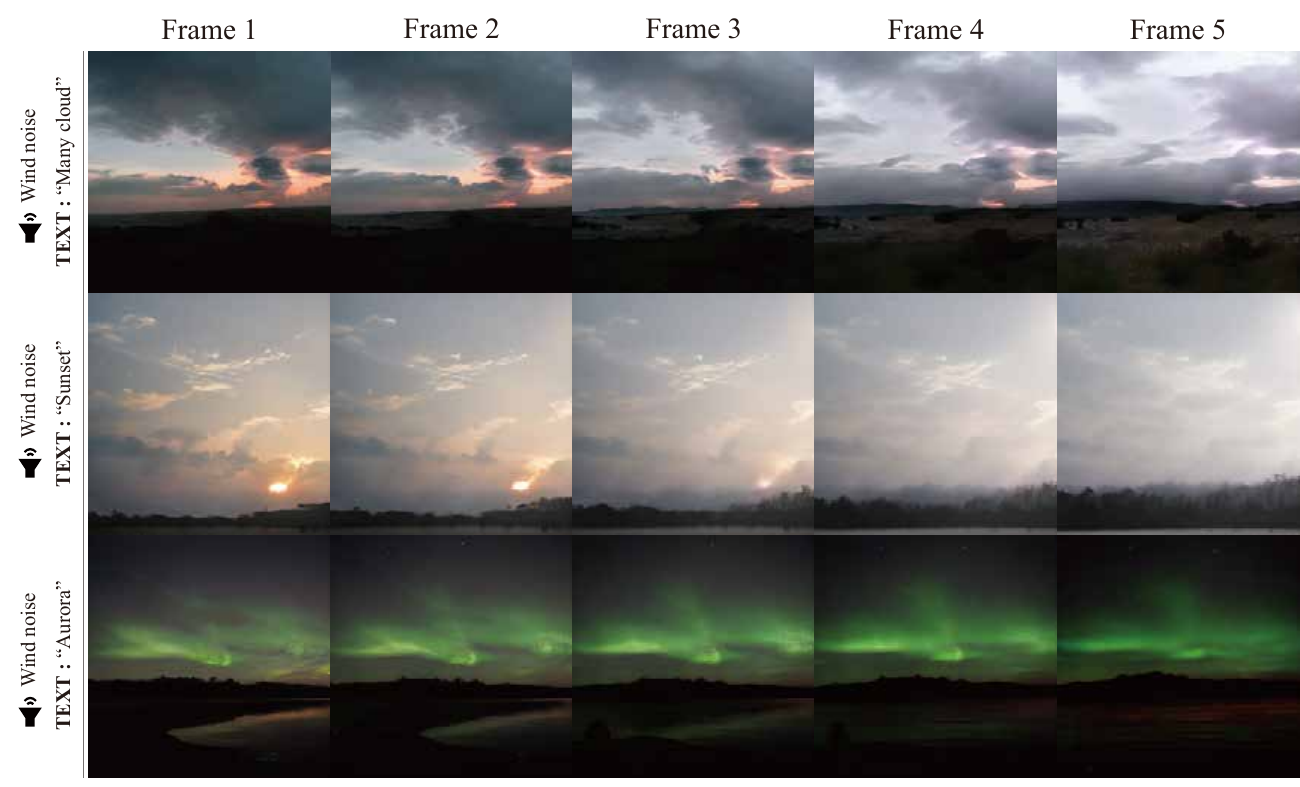}
    \caption{Examples of sound-guided semantic video generation from sound with text prompts. Our model generates the videos by guiding sound from latent code containing text content. The time interval between each column is five frames.}
    \label{fig:sunset-figure}
\end{figure}

\subsection{Sound-guided Video Editing with Text Constraints}
The combination of text and sound makes it possible to generate various videos with rich information. Fig.~\ref{fig:sunset-figure} demonstrates that the combination of text-query constraints and sound results in a much more versatile style of video. Because our model shares the CLIP~\cite{radford2learning} space trained with large-scale text and image pairs, video generation with text constraint is available. In addition, it is possible to apply styles of different levels with the style mixing technique.
For video editing, we randomly sample multiple latent codes. The initial latent code for generating the first frame is the sample with the closest CLIP embedding cosine distance of the text prompt.

\section{Discussion}
\myparagraph{Limitations.}
In our method, we sample the first frame from the StyleGAN~\cite{karras2021alias,karras2019style} latent space, and the space covers only the train domain. So, color change is rarely observed when the pre-trained StyleGAN and the video datasets have a domain gap. Still, we observe that our model is applicable to many cases such as sound-guided face video generation~(see sup.).

In addition, in order for our model to learn to generate a video, the weights of the pre-trained image generator are required, which increases the training time. And maintaining the movement size and identity of the latent code predicted by the recurrent block has a tradeoff relationship~(see sup.). 

\myparagraph{Societal Impact.} Following this characteristic of our method, the user can generate a real video with the user's desire, and this usefulness allows the user to see the video that existed only in the user's imagination. However, although these features can provide a suitable output video to the user if the input video source is a work of art or is not ethically correct, the result produced may imply the contrary intentions of the original video creator or may raise ethical concerns. 

\section{Conclusion}
This paper shows a new way to make sound-guided realistic videos by taking advantage of the multi-modal embedding space, which uses sound, images, and text.
To be more specific, we use CLIP~\cite{radford2learning} space to encode the sound information into the StyleGAN~\cite{karras2021alias,karras2019style} latent space, which allows generating videos that contain the corresponding source sound semantics.
We design our model with the recurrent neural network since video consists of multiple frames and requires time-domain consistency.
Multiple recurrent blocks temporarily represent rich style information~(view point, sound-dependent frame's fine features).
Additionally, we curate a new high-fidelity audio-video landscape dataset to validate our proposed method, which is superior to other methods.
We demonstrate that our method qualitatively and quantitatively outperformed state-of-the-art methods for sound-guided video generation. 
The proposed model can be used in various applications such as video generation with text constraints. 




\newpage

\clearpage
%
%
\bibliographystyle{splncs04}
\bibliography{egbib}
\end{document}